\documentclass[Afour,sageh,times]{sagej}

\usepackage{moreverb,url}
\usepackage{caption}
\usepackage{subcaption}
\usepackage{dirtree}
\usepackage{tabularx}

\newcolumntype{Y}{>{\centering\arraybackslash}X}
\newcommand{\best}[1]{{\bf\color{blue}#1}}

\usepackage[colorlinks,bookmarksopen,bookmarksnumbered,citecolor=red,urlcolor=red]{hyperref}

\newcommand\BibTeX{{\rmfamily B\kern-.05em \textsc{i\kern-.025em b}\kern-.08em
T\kern-.1667em\lower.7ex\hbox{E}\kern-.125emX}}

\begin{document}

\runninghead{Boittiaux \textit{et al.}}

\title{Eiffel Tower: A Deep-Sea Underwater Dataset for Long-Term Visual Localization}

\author{Cl\'ementin Boittiaux\affilnum{1,2,3},
Claire Dune\affilnum{2},
Maxime Ferrera\affilnum{1},
Aur\'elien Arnaubec\affilnum{1},
Ricard Marxer\affilnum{3},
Marjolaine Matabos\affilnum{4},
Lo\"ic Van Audenhaege\affilnum{4} and
Vincent Hugel\affilnum{2}}

\affiliation{\affilnum{1}Ifremer, Zone Portuaire de Br\'egaillon, La Seyne-sur-Mer, France\\
\affilnum{2}Universit\'e de Toulon, COSMER, Toulon, France\\
\affilnum{3}Universit\'e de Toulon, Aix Marseille Univ, CNRS, LIS, Toulon, France\\
\affilnum{4}Univ Brest, CNRS, Ifremer, UMR6197 BEEP, F-29280 Plouzan\'e, France}

\corrauth{Cl\'ementin Boittiaux, Ifremer Centre M\'editerran\'ee,
Zone Portuaire de Br\'egaillon,
83500 La Seyne-sur-Mer, France}

\email{boittiauxclementin@gmail.com}

\begin{abstract}
Visual localization plays an important role in the positioning and navigation of robotics systems within previously visited environments. When visits occur over long periods of time, changes in the environment related to seasons or day-night cycles present a major challenge. Under water, the sources of variability are due to other factors such as water conditions or growth of marine organisms. Yet it remains a major obstacle and a much less studied one, partly due to the lack of data. This paper presents a new deep-sea dataset to benchmark underwater long-term visual localization. The dataset is composed of images from four visits to the same hydrothermal vent edifice over the course of five years. Camera poses and a common geometry of the scene were estimated using navigation data and Structure-from-Motion. This serves as a reference when evaluating visual localization techniques. An analysis of the data provides insights about the major changes observed throughout the years. Furthermore, several well-established visual localization methods are evaluated on the dataset, showing there is still room for improvement in underwater long-term visual localization. The data is made publicly available at \href{https://www.seanoe.org/data/00810/92226/}{seanoe.org/data/00810/92226/}.
\end{abstract}


\maketitle

\section{Introduction}

With the advent of Autonomous Underwater Vehicles (AUVs) and Remotely Operated Vehicles (ROVs), there is a need to enable these vehicles to localize themselves accurately in an underwater environment. This paper addresses the problem of visual localization which consists in estimating the 6 degrees-of-freedom (6DOF) pose of a camera given its image and previous observations made in the area. It has received a lot of attention over the last decade with the rise of self-driving cars. This task becomes more difficult when the environment is subject to important changes, as it is the case for images acquired during successive visits in deep ocean. Long-term localization methods aim to deal with major changes in the environment, \textit{e.g.}, snow during winter \citep{sattler2018benchmarking}.

\begin{figure}[t]
  \centering
  \includegraphics[width=\columnwidth]{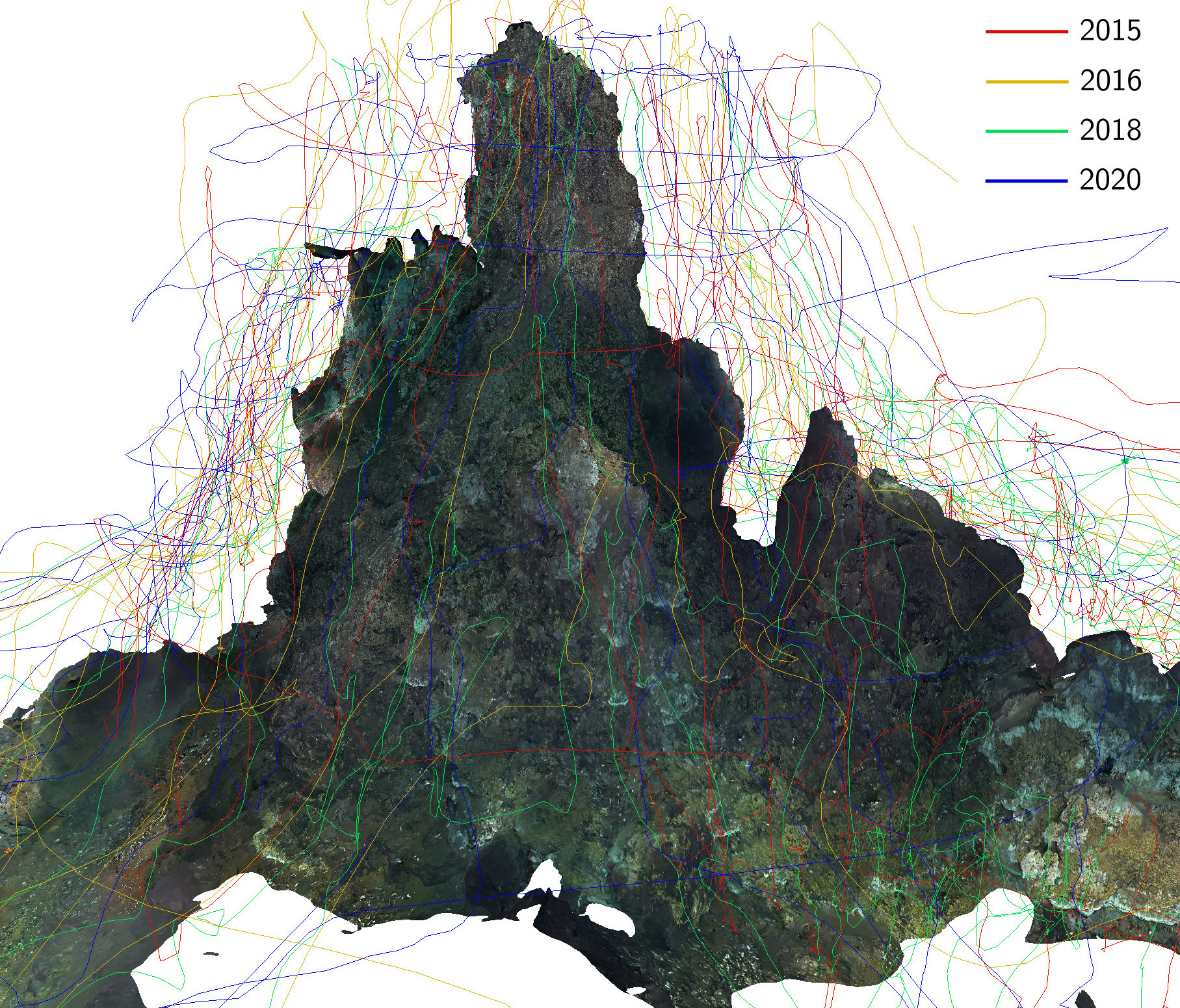}
  \caption{Trajectories along the \textit{Eiffel Tower} hydrothermal vent. Camera poses were retrieved using COLMAP \citep{schoenberger2016sfm}. 3D model was meshed and textured using OpenMVS \citep{openmvs2020}.\label{fig:traj}}
  \vspace{-20pt}
\end{figure}

Most of the databases used to evaluate these methods are made up of terrestrial data. They cover a wide range of environmental changes such as day-night, seasons and weather conditions \citep{griffith2017symphony,sattler2018benchmarking,burnett_boreas22}. 

Because underwater images have different sources of technical and environmental variability, existing datasets are not suitable for evaluating long-term localization performance in such scenarios. Indeed, the characteristics of the underwater medium cause many visual perturbations related to light and color absorption, turbidity and back-scattering. Furthermore, in deep-sea scenarios, underwater vehicles must be equipped with artificial lighting systems in order to illuminate the absolute darkness of the environment.  While this allows the use of cameras to record what lies on the seafloor, it also creates strong differences in illumination depending on the distance between the robot and the scene. In addition to these short-term perturbations, long-term changes occur in these environments, \textit{e.g.}, changes in the distribution of microbial and animal communities or topographical changes.

Some localization methods rely on deep retrieval and feature matching networks to be robust to large sources of variability. However, these networks, \textit{e.g.}, NetVLAD \citep{netvlad}, have only been pre-trained on terrestrial data. Applications of these models in underwater environments may not be straightforward due to the domain shift associated to the specificity of underwater imaging.

Previously published underwater datasets targeting online localization \citep{aqualoc,mallios2017sonar} span over very short temporal ranges, \textit{i.e.}, data were acquired during the same day. Thus, they do not cover the long-term changes that can appear in these environments, leaving a gap in methods and datasets available to treat multiannual deep-sea image sequences. 

This paper presents a new dataset for long-term visual localization in a deep-sea environment. It is composed of four different visits of the same hydrothermal edifice over five years \citep{GIRARD2020102397} (Figure~\ref{fig:traj}). More specifically, it provides the following data:
\begin{itemize}
    \item Images of the vent for all four visits.
    \item Navigation data in the form of latitude, longitude and altitude information.
    \item 3D models of the scene estimated using Structure-from-Motion (SfM) for each visit year.
    \item A global 3D model including all images in a common reference frame.
\end{itemize}
The data presents changes over time related to all the aforementioned underwater imaging factors. It also presents some peculiar characteristics, like the occurrence of black and white smokers that emit hot hydrothermal fluid. Moreover, due to the numerous sources of variation present in this dataset, it may also be used for detecting long-term changes that take place in deep-sea environments. This paper makes the following contributions: i) it provides a new publicly available dataset for long-term visual localization in a deep-sea environment; ii) it presents an analysis of environmental and topographic changes between the different visits; iii) it benchmarks several visual localization methods on the given dataset.

\section{Related work}

Datasets used for benchmarking visual localization algorithms are mostly terrestrial, including Aachen Day-Night, RobotCar Seasons and CMU Seasons introduced by \cite{sattler2018benchmarking} as well as Cambridge \citep{kendall2015posenet}, 7-Scenes \citep{shotton2013_7scenes} and 12-Scenes \citep{valentin2016_12scenes}. 7-Scenes and 12-Scenes are collected in an indoor setting, while the others are composed of outdoor environments. \cite{sattler2018benchmarking} datasets exhibit some difficult localization scenarios, like day-night observations. Similar datasets to benchmark the visual localization task under water are rare due to the cost of data collection. 

Existing underwater datasets \citep{mallios2017sonar,aqualoc} focus on providing data for the development of underwater SLAM algorithms. AQUALOC dataset \citep{aqualoc} provides underwater monochromatic images synchronized with inertial and depth data for three different sites off Corsica. One of the sites is a harbor lying at a depth of 3 to 4 m and the other two are archaeological sites that lie at depths of 270 m and 380 m. While sequences follow different trajectories, all different visits occurred during the same day, not covering all the possible changes that can happen in this environment, \textit{e.g.}, salinity variation that can alter the pinhole model, increased turbidity, sedimentation or marine population changes. \cite{posenet_ntnu} evaluated PoseNet \citep{kendall2015posenet}, an end-to-end visual localization neural network, on an underwater dataset acquired in a pool where camera poses were obtained with an underwater motion capture system. Other underwater datasets that focus on different tasks, \textit{e.g.}, dehazing \citep{seathru,UIEB_dataset,Berman_restoration}, do not provide images of the same site over long periods of time.

\cite{toureiffel_surface} already published image data of the 2015 visit of the \textit{Eiffel Tower} vent. The authors presented a novel method for surface reconstruction of underwater structures. To benchmark their algorithm, they used a point cloud resulting from a SfM on the images of the 2015 dive. The current work also exploits SfM to produce the reference poses and scene geometry on this collection, and completes it with similar data from three other dives of the same site in different years, capturing long-term changes.

\begin{figure}
    \centering
    \includegraphics[width=\columnwidth]{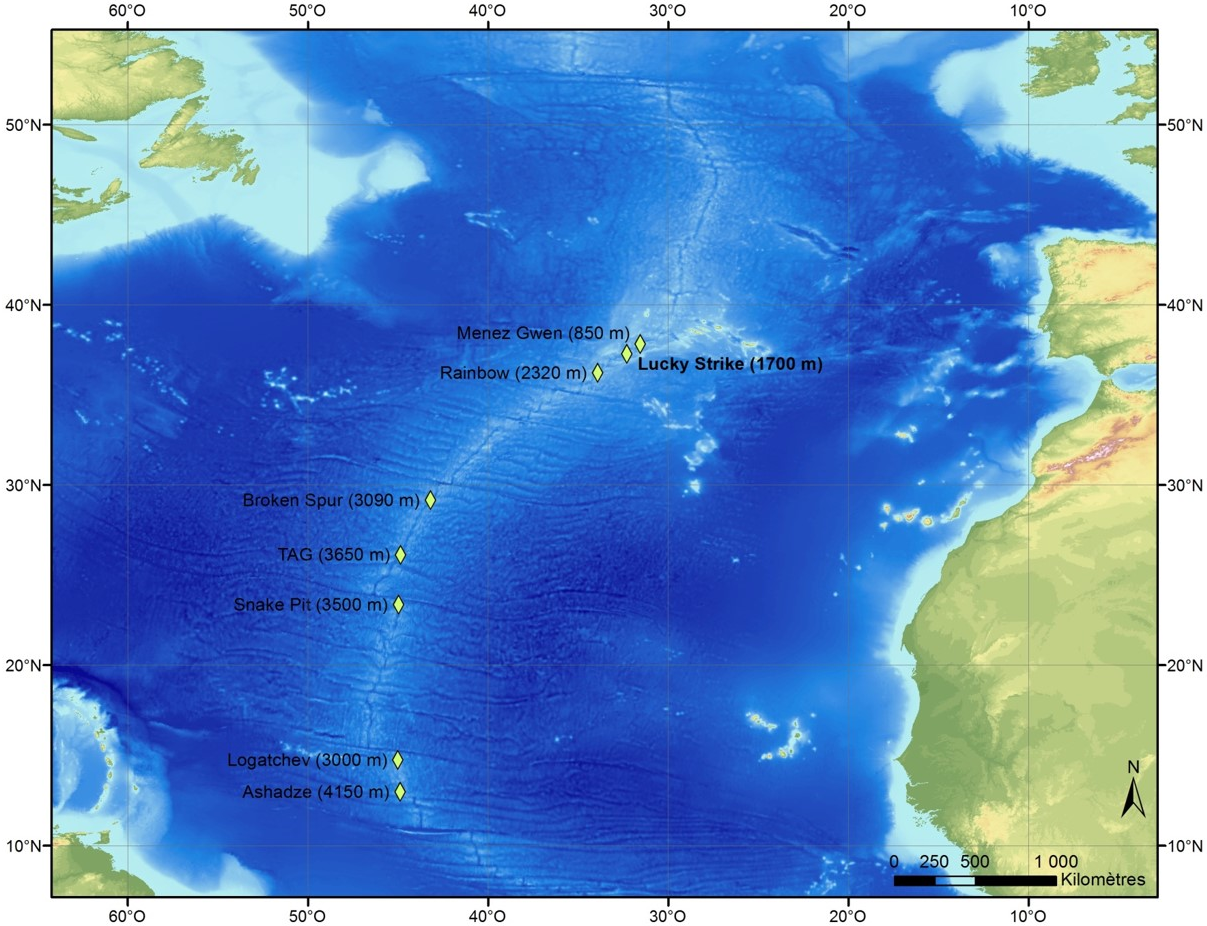}
    \caption{Location of the Lucky Strike vent field on the Mid-Atlantic Ridge (Sources: Esri, GEBCO, NOAA, National Geographic, DeLorme, HERE, Geonames.org).\label{fig:lucky}}
\end{figure}

\begin{figure*}
  \centering
  \begin{subfigure}[b]{\textwidth}
      \centering
      \includegraphics[width=\textwidth]{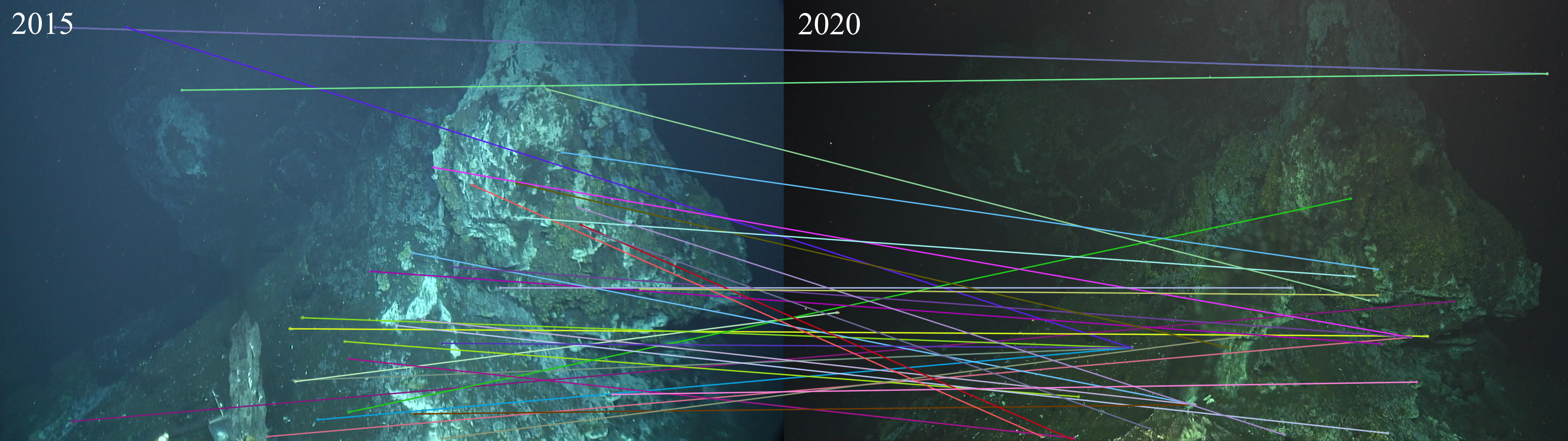}
      \caption{SIFT using brute-force matching.}
      \vspace{8pt}
  \end{subfigure}
  \begin{subfigure}[b]{\textwidth}
      \centering
      \includegraphics[width=\textwidth]{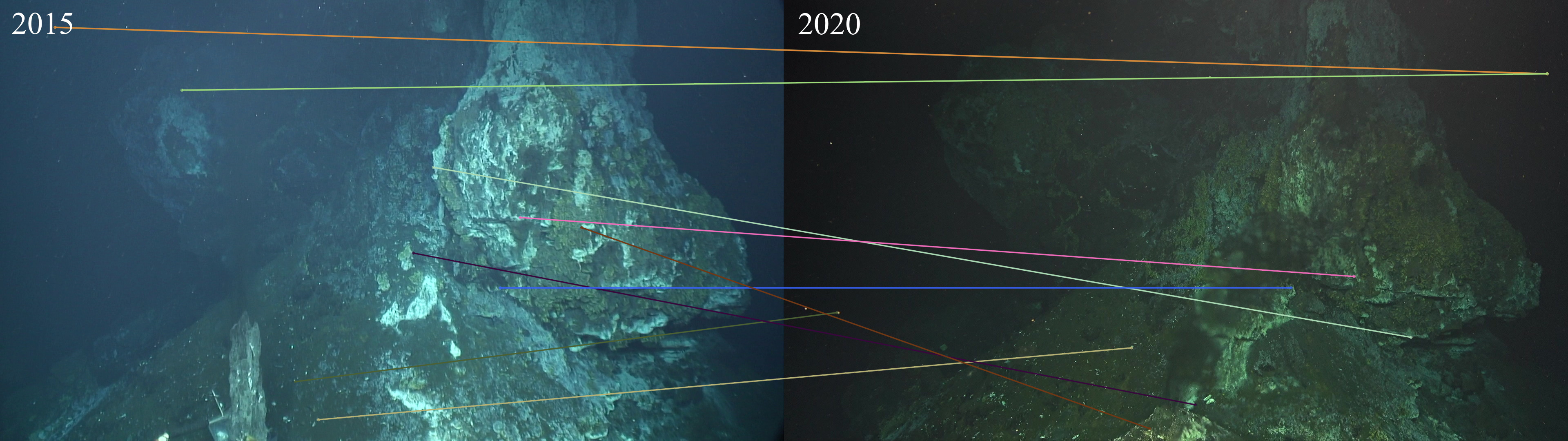}
      \caption{SIFT using brute-force matching. Matches are then filtered after estimating the fundamental matrix within a RANSAC scheme.}
      \vspace{8pt}
  \end{subfigure}
  \hfill
  \begin{subfigure}[b]{\textwidth}
      \centering
      \includegraphics[width=\textwidth]{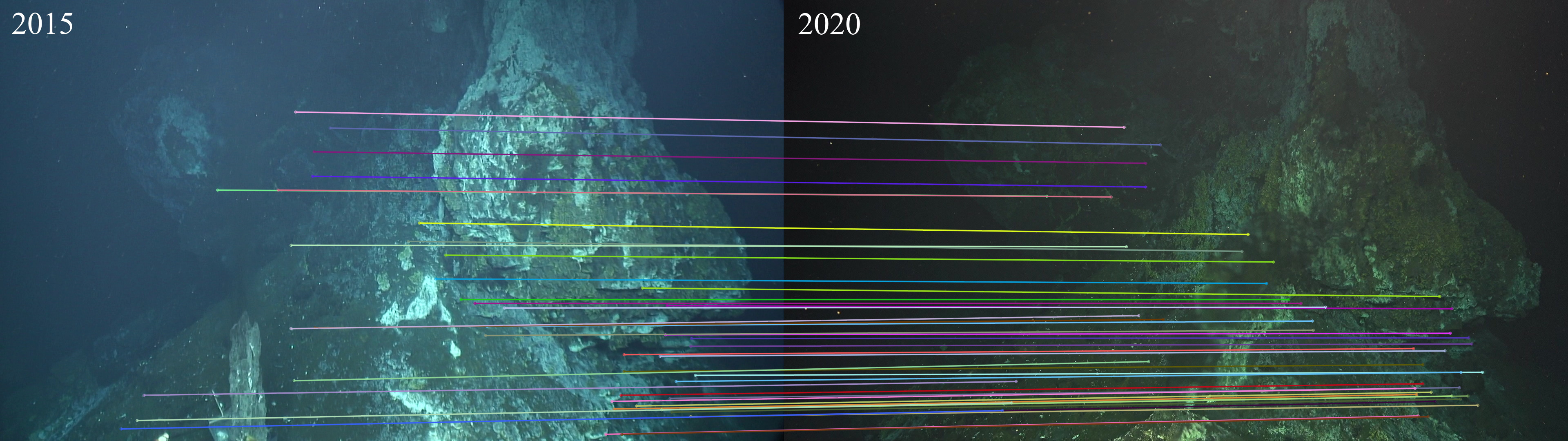}
      \caption{SuperPoint \citep{superpoint} followed by SuperGlue \citep{sarlin20superglue}.}
  \end{subfigure}
     \caption{Feature matching between cross-year images using different methods.}
     \label{fig:matches}
\end{figure*}

Visual localization benchmarking datasets require reference camera poses for each image, which can be constructed in different ways. Most common methods to access such information as well as the scene's geometry rely on SfM or depth-based SLAM \citep{brachmann2021limits}. In most deep-sea missions, the use of motion capture is discarded due to the difficulties in deploying such a system.
Standard RGB-D sensors cannot directly be used underwater because of the absorption of infrared light in water, making depth-based SLAM harder to set up. Thereby, SfM offers a practical solution for estimating camera poses and scene geometry in the underwater environment. Nevertheless, \cite{brachmann2021limits} showed that the performance of a localization method on a given dataset is greatly affected by the method used to build the ``ground truth'' of this dataset. Indeed, methods that minimize the same error as the algorithm used for estimating the ground truth poses have the advantage of leading to the same local minima. For this reason, methods that favor SfM-based ground truths may perform better on our dataset.

\begin{figure*}
    \centering
    \includegraphics[width=\textwidth]{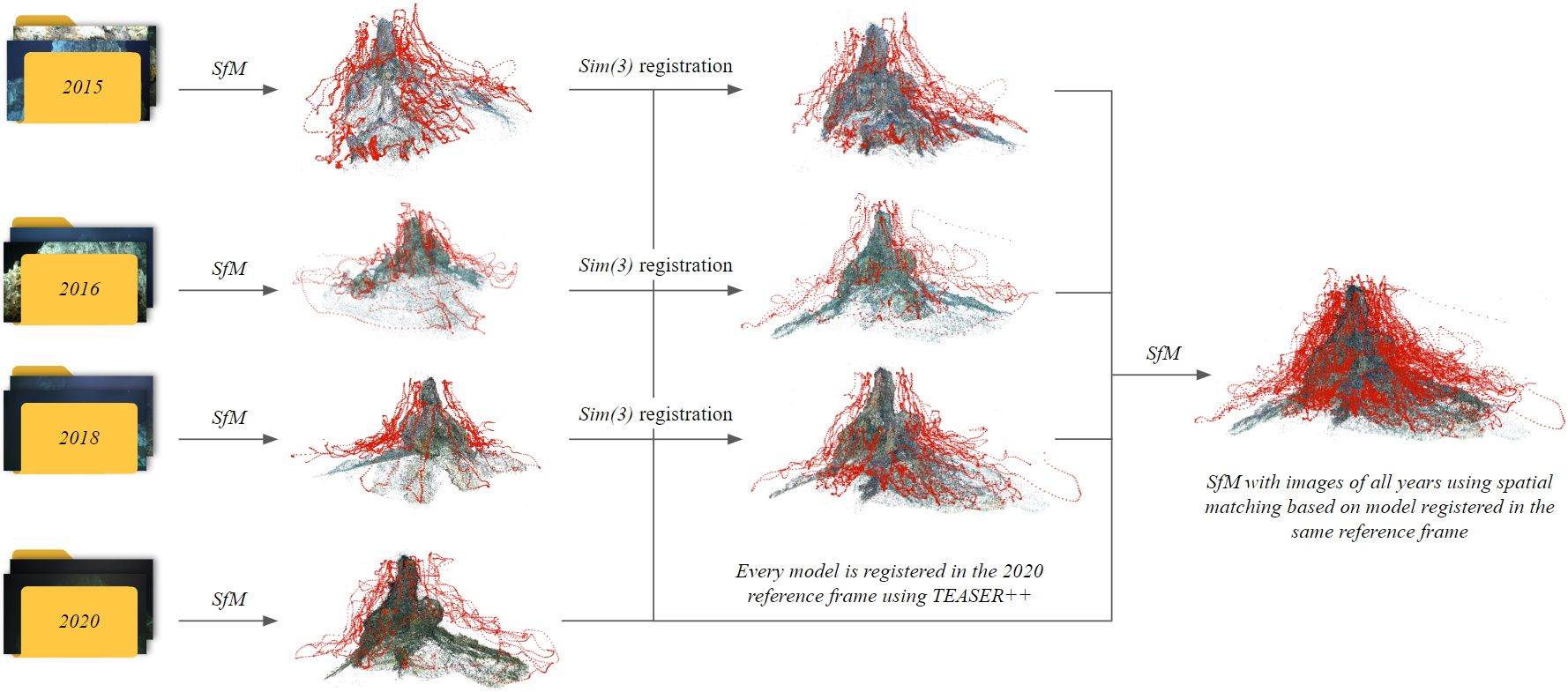}
    \caption{Structure-from-Motion pipeline to match images across years. Models are first built independently for each year. They are then registered in a common reference frame, \textit{i.e.}, 2020 reference frame, using TEASER++ \citep{teaserpp} and ICP \citep{Zhou2018}. Finally, a model embedding images of all years is computed using spatial matching based on the camera poses of individual models that now share a common reference frame.\label{fig:sfmpip}}
\end{figure*}

\section{Data collection}

The EMSO-Azores deep-sea observatory on the Mid-Atlantic Ridge supports the long-term monitoring of the \textit{Lucky Strike} hydrothermal vent field (Figure~\ref{fig:lucky}) since 2010. During the annual maintenance cruises \citep{momarsat}, a ROV operated by the French National Institute for Ocean Science (Ifremer) has been used to study the evolution of the hydrothermal circulation and the associated fauna communities over several years \citep{matabos_integrating_2022}. Within this field, the hydrothermal vent edifice \textit{Eiffel Tower}, located at 1700 m beneath the surface, has been extensively studied since its discovery in 1992 \citep{langmuir}. Four dives, in 2015, 2016, 2018 and 2020, were dedicated to the 3D reconstruction of the structure enabling quantitative monitoring of vent community distribution and dynamics \citep{GIRARD2020102397}.

\begin{table}
    \small\sf\centering
    \caption{Cameras settings.\label{table:cameras}}
    \begin{tabularx}{\linewidth}{XYY}
    \toprule
    Year&Resolution&Frame rate\\
    \midrule
    2015&1920x1080 px&25 fps\\
    2016&1920x1080 px&25 fps\\
    2018&1920x1080 px&25 fps\\
    2020&3840x2160 px&30 fps\\
    \bottomrule
    \end{tabularx}
\end{table}

\begin{table*}[t]
    \small\sf\centering
    \caption{Reconstruction statistics. For each model, we report the number of registered images, the number of triangulated 3D points, the mean track length (number of images in which a 3D point is observed), the mean number of 2D observations per image as well as the mean reprojection error in pixels.\label{tab:stats}}
    \begin{tabularx}{\linewidth}{XYYYcY}
    \toprule
    Model & Nb. of images & Nb. of 3D points & Mean track length & Mean obs. per image & Mean reproj. error\\
    \midrule
    2015 & 4,914 & 525,522 & 8.48 & 906.4 & 1.35 px\\
    2016 & 3,699 & 520,320 & 5.85 & 823.5 & 1.32 px\\
    2018 & 5,493 & 618,421 & 7.09 & 798.1 & 1.31 px\\
    2020 & 3,976 & 464,331 & 8.35 & 975.5 & 1.33 px\\
    \cmidrule(lr){1-6}
    Global & 18,082 & 1,971,726 & 8.24 & 898.7 & 1.39 px\\
    \bottomrule
    \end{tabularx}
\end{table*}

During the different dives, synchronized videos and navigation data were acquired using the ROV sensors. Videos were captured using two different cameras, whose characteristics are presented in Table~\ref{table:cameras}.
Images were acquired through a specifically designed dome port with corrective lenses to account for underwater refraction. As these lenses largely compensate the distortion induced by the air-glass-water mediums, a second order radial distortion model with k1 and k2 distortion coefficient was used to calibrate the cameras underwater. The ROV incorporated an artificial lighting system consisting of 12 LED panels delivering 20,000 lumens each. It also embedded an Ultra-Short Baseline acoustic positioning system (USBL), an Inertial Navigation System (INS), a Doppler Velocity Log (DVL) and a depth sensor that were fused similarly to \cite{guerrero2016navigation} to compute the navigation data which provide an estimate of the position of the vehicle. This position is composed of the latitude, longitude and altitude of the ROV as well as its yaw, pitch and roll angles. However, navigation data are only consistent within each visit due to the uncertainty of the USBL, which can exhibit offsets of several meters between the frames of the different visits.

\section{Camera poses and scene geometry}

This visual localization dataset offers images and their 6DOF poses expressed in a common frame of reference. COLMAP SfM \citep{schoenberger2016sfm} was used to obtain camera poses and intrinsic parameters, as well as 3D scene geometry.

From the videos captured each year, one image was extracted every 3 seconds to create the input data of the dataset. Some images were polluted on their border with small navigation overlays that were removed using an inpainting technique \citep{telea}. In addition, images were inspected manually to discard irrelevant ones, \textit{e.g.}, images only capturing the water column. 

Several issues were encountered when attempting to use COLMAP directly on images from all years. Firstly, to create image pairs, COLMAP relies on image retrieval methods like vocabulary tree \citep{schoenberger2016vote} or NetVLAD \citep{netvlad}. While these methods successfully match images of the same year, they show poor performance at pairing images across different years. 
Secondly, to perform feature matching between image pairs, COLMAP uses SIFT descriptors. However, as shown on Figure~\ref{fig:matches}, hand-crafted descriptors fail to produce satisfactory matches between images of different years.

To overcome these issues, we adopted the SfM pipeline described on Figure~\ref{fig:sfmpip}\footnote{The code used to compute the different steps of the presented SfM pipeline is available at \href{https://github.com/clementinboittiaux/sfm-pipeline}{github.com/clementinboittiaux/sfm-pipeline}.}. We first built a model for each year independently. For each individual model, image retrieval was performed using navigation data. However, navigation data are missing for 3,178 images of the 2015 dive. In this case, image retrieval was achieved using NetVLAD \citep{netvlad}. Within the same visit, NetVLAD proved to be efficient to retrieve similar images. Instead of hand-crafted descriptors, we used SuperPoint \citep{superpoint} and SuperGlue \citep{sarlin20superglue} for feature matching. We then registered the resulting 3D point clouds of each individual model to the 2020 point cloud using TEASER++ \citep{teaserpp} and refined the result with ICP \citep{Zhou2018}. This way, we obtained a coarse estimation of the camera intrinsics and poses of each year in the same reference frame. We then paired cross-year images using these poses and matched these images with SuperPoint and SuperGlue. Finally, we used COLMAP to obtain a global model embedding all images.

The scale of each model was retrieved by aligning camera poses in $\mathit{Sim}(3)$ with available navigation data using Umeyama's algorithm \citep{umeyama}. For the global model that includes all observations, the alignment was performed using only 2020 navigation data. At this point, we have the 6DOF pose of each viewpoint as well as a 3D point cloud of the full vent in the same reference frame at real scale for every year.


\begin{table}
    \small\sf\centering
    \caption{Percentage of paired features indexed by the years in which the two images are taken. The table is normalized so that rows add up to 100\%. It shows the amount of cross-years coverage of observations used to perform the SfM.
    \label{tab:shared}}
    \begin{tabularx}{\linewidth}{Xcccc}
    \toprule
    Observation year&2015&2016&2018&2020\\
    \midrule
    2015&65.9\%&10.7\%&15.2\%&8.12\%\\
    2016&20.3\%&50.9\%&16.2\%&12.6\%\\
    2018&16.1\%&9.05\%&63.1\%&11.8\%\\
    2020&9.79\%&8.00\%&13.5\%&68.8\%\\
    \bottomrule
    \end{tabularx}
\end{table}

Table \ref{tab:stats} reports statistics about the reconstructions obtained with COLMAP as a way to illustrate the certainty level of the proposed ground truth.
Table \ref{tab:shared} reports the percentage of 3D points matched between each year in the resulting SfM. While the majority of 3D points observations are contained within the same year, a significant proportion of them were matched across different years. This ensures that the scene geometry and camera poses are consistent between the different visits.

\section{Dataset format}
The \textit{Eiffel Tower} dataset is composed of images of the dives and a global COLMAP model embedding all visits. For reproducibility purposes, we also provide individual COLMAP models for each visit and interpolated navigation data for each image when available. The dataset architecture is detailed in Figure~\ref{fig:arch}.

Each image is named after the date at which it was acquired in the format \verb+YYYYmmddTHHMMSS.fffZ+, where \verb"YYYY" is the four digit year, \verb"mm" is the month, \verb"dd" is the day, \verb"HH" is the hour in the 24 hours format, \verb"MM" are the minutes, \verb"SS" the seconds and \verb"fff" the milliseconds.
A COLMAP model embeds scene geometry, camera intrinsics and poses. It is composed of 3 files: \verb"cameras.txt" contains the camera intrinsics; \verb"images.txt" provides camera poses as well as lists of 2D keypoints associated to their 3D observations for each image; \verb"points3D.txt" consists of the positions and colors of the 3D points. The specificities of COLMAP output format can be found in the documentation\footnote{\href{https://colmap.github.io/format.html}{colmap.github.io/format.html}}. Moreover, COLMAP provides scripts to easily read models in C++, Python and MATLAB\footnote{\href{https://github.com/colmap/colmap/blob/dev/scripts}{github.com/colmap/colmap/blob/dev/scripts}}.
Navigation data are provided in the following text format:\\
\begin{tabular}{llll}
    \textit{image001.png}&\textit{lat}$_\mathit{1}$&\textit{lon}$_\mathit{1}$&\textit{alt}$_\mathit{1}$\\
    \textit{image002.png}&\textit{lat}$_\mathit{2}$&\textit{lon}$_\mathit{2}$&\textit{alt}$_\mathit{2}$\\
    \textit{...}
\end{tabular}\\
where \textit{lat}, \textit{lon} and \textit{alt} are the latitude, longitude and altitude of the vehicle.


\begin{figure}
\dirtree{%
.1 EiffelTower.
.2 global.
.3 sfm.
.4 cameras.txt.
.4 images.txt.
.4 points3D.txt.
.2 2015.
.3 images.
.3 sfm.
.3 navigation.txt.
.2 2016.
.2 2018.
.2 2020.
}
\caption{Dataset file organization. For each year, images, navigation data and an individual SfM are provided. A global SfM embedding all years for benchmarking visual localization methods is also available.\label{fig:arch}}
\end{figure}

\section{Characterizing changes across years}

\begin{figure*}
  \centering
  \includegraphics[width=\textwidth]{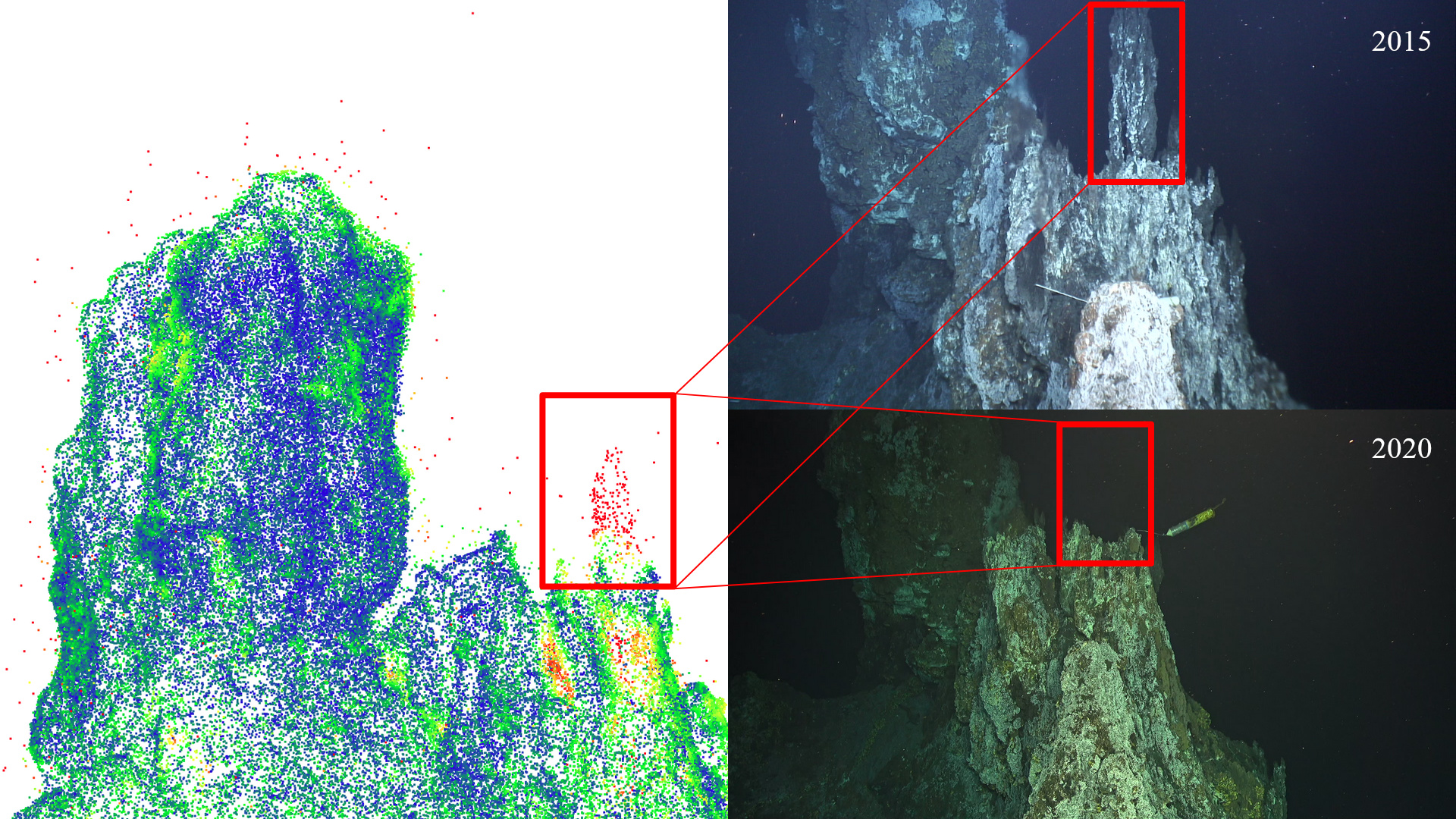}
  \caption{Illustration of a topological modification. The left image shows a point cloud distance between 2015 and 2020 models. We notice a piece from 2015 missing in 2020. This modification is visible on the right images.\label{fig:topo}}
\end{figure*}

\begin{figure*}
  \centering
  \includegraphics[width=\textwidth]{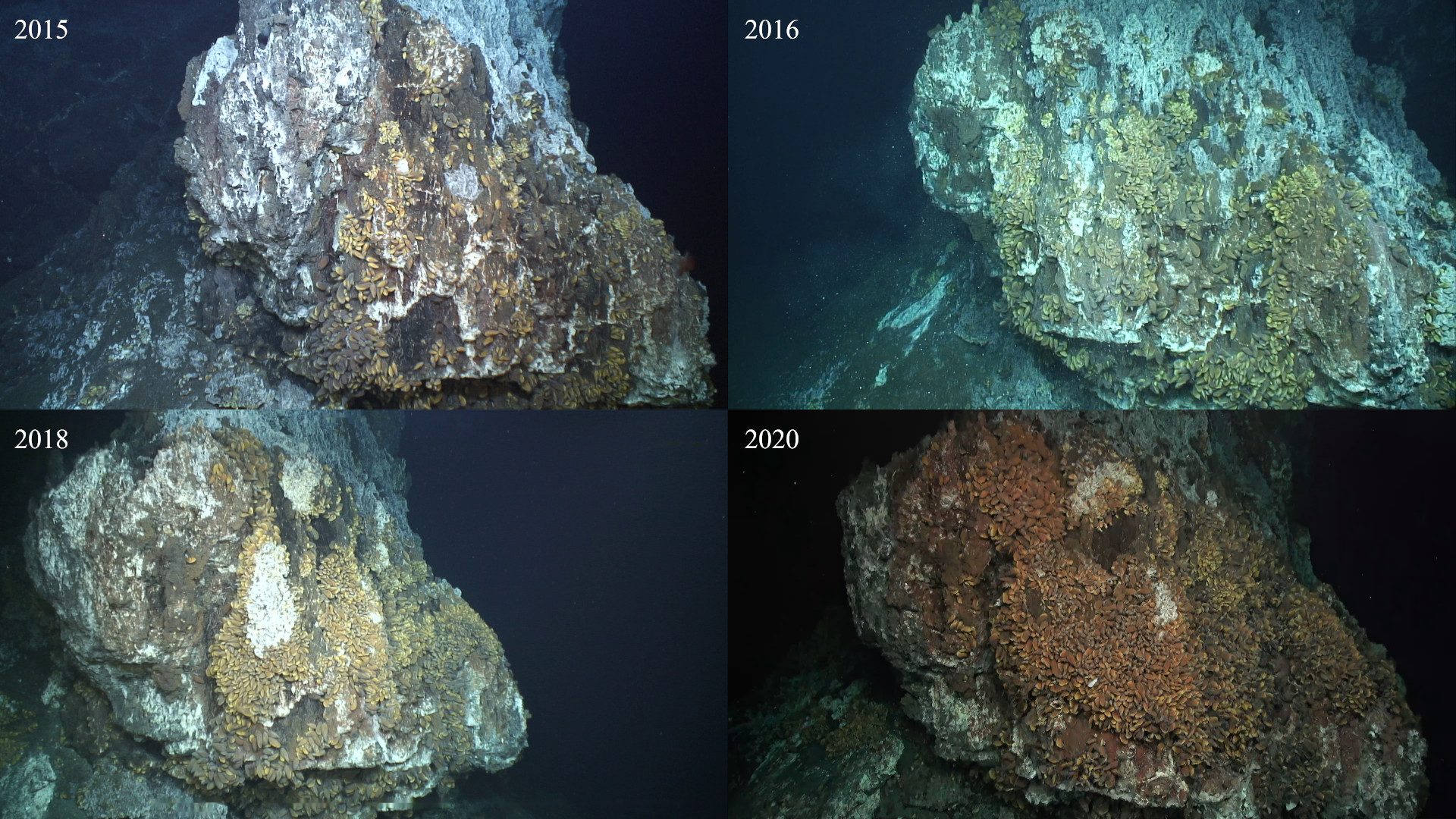}
  \caption{Evolution of the south-east fa\c cade of the vent. A growth in the mussels' population significantly alters the visual aspect of the scene, making it difficult to match specific 2D points.\label{fig:changes}}
\end{figure*}

\begin{figure*}
  \centering
  \begin{subfigure}[b]{0.49\textwidth}
      \centering
      \includegraphics[width=\textwidth]{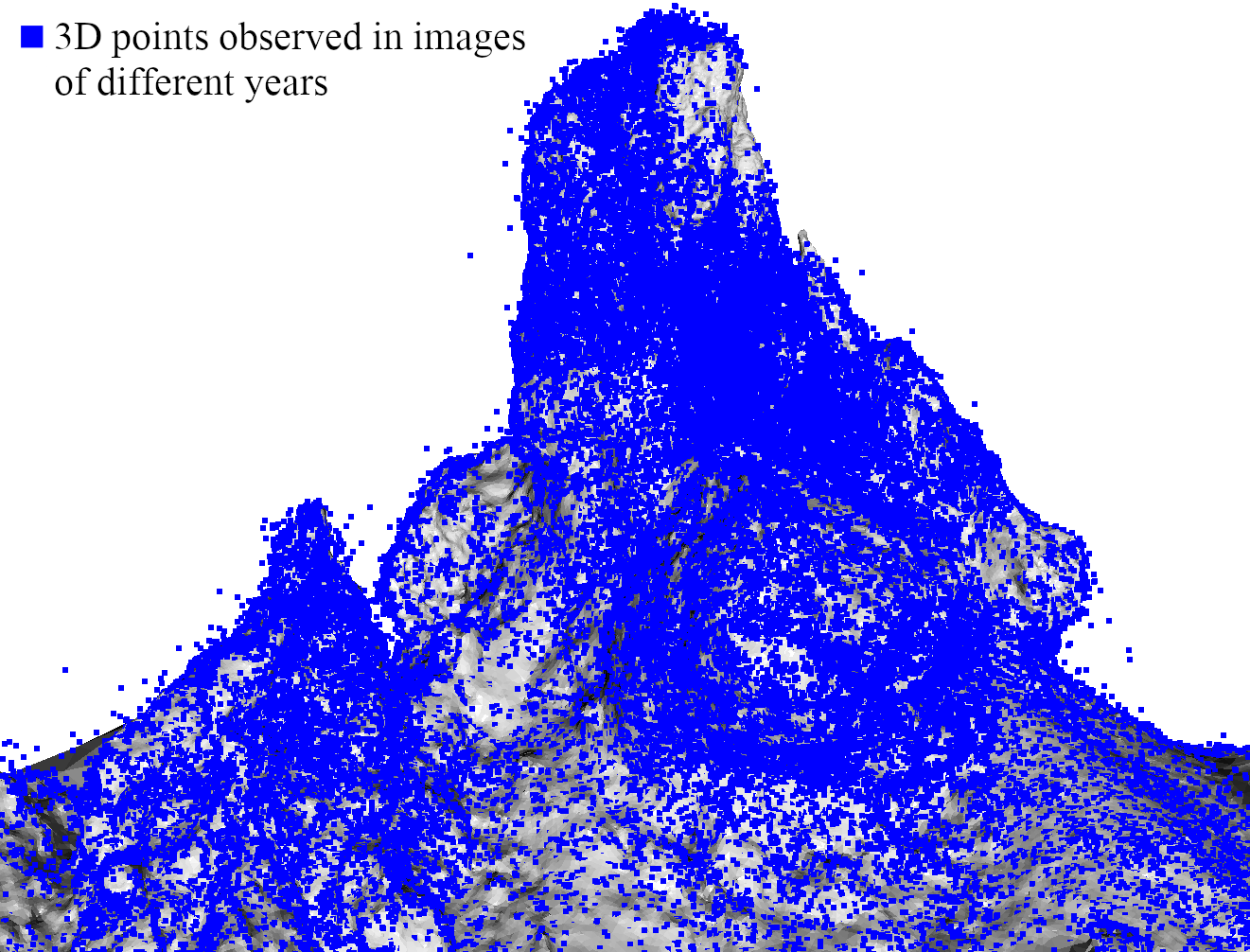}
      \caption{West fa\c cade.}
  \end{subfigure}
  \hfill
  \begin{subfigure}[b]{0.49\textwidth}
      \centering
      \includegraphics[width=\textwidth]{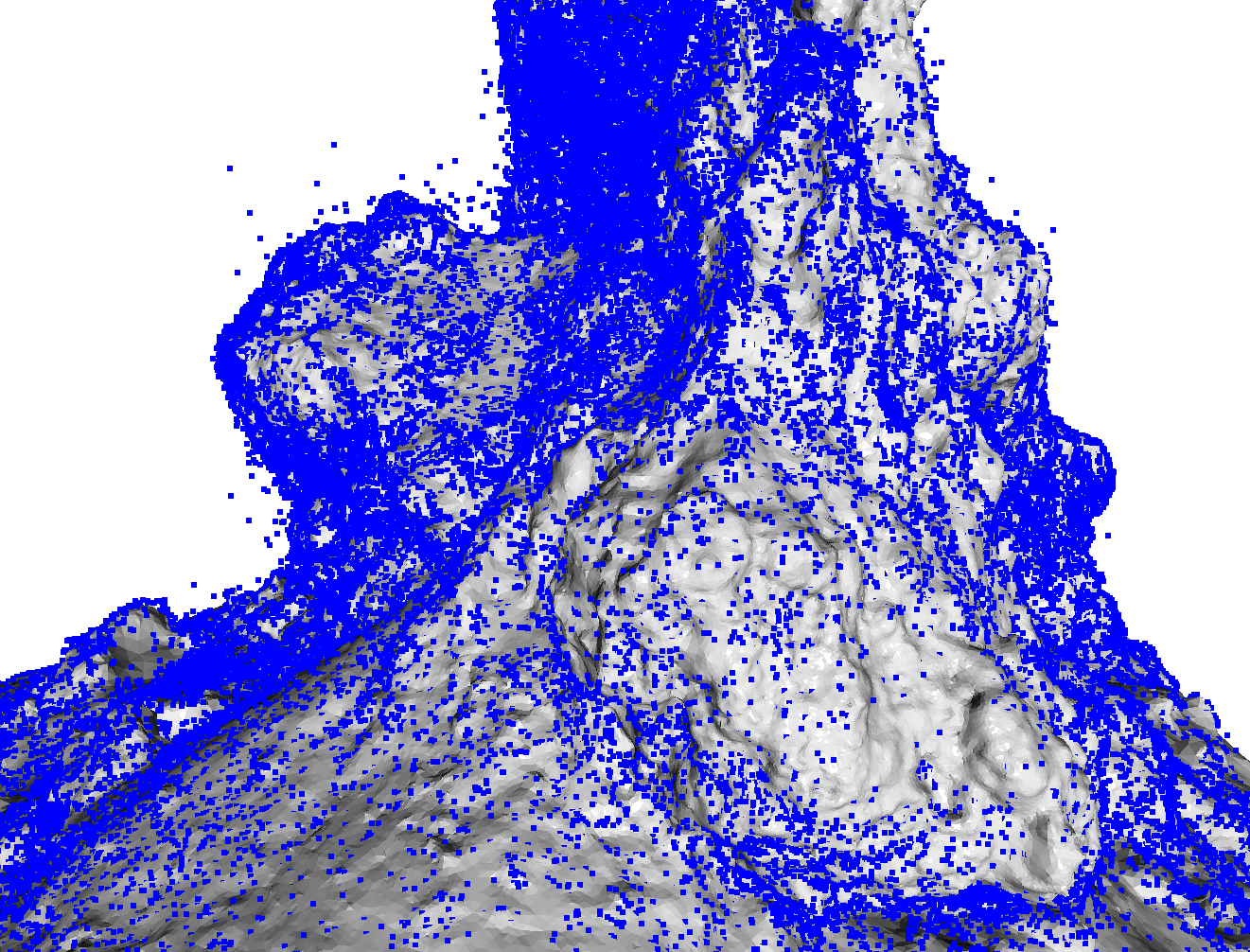}
      \caption{South-east fa\c cade.}
  \end{subfigure}
  \caption{Distribution of 3D points that are triangulated between images of different years on the Eiffel Tower edifice. 3D points resulting from cross-years triangulation are more scarce on the south-east fa\c cade due to biological changes.}
  \label{fig:cross-year-matching}
\end{figure*}

\begin{figure*}
  \centering
  \begin{subfigure}[b]{0.3\textwidth}
      \centering
      \includegraphics[width=\textwidth]{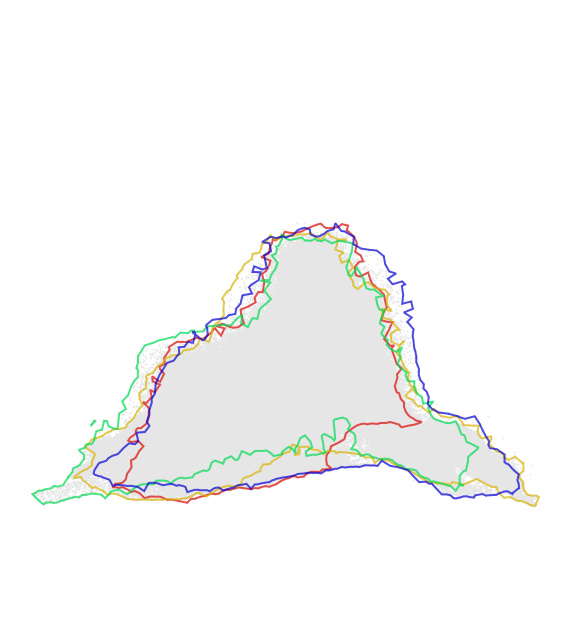}
      \caption{East-Up View.}
  \end{subfigure}
  \hfill
  \begin{subfigure}[b]{0.3\textwidth}
      \centering
      \includegraphics[width=\textwidth]{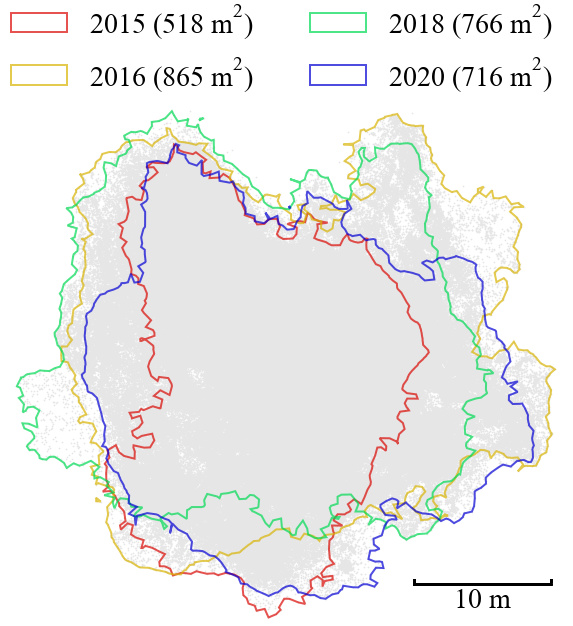}
      \caption{East-North View.}
  \end{subfigure}
  \hfill
  \begin{subfigure}[b]{0.3\textwidth}
      \centering
      \includegraphics[width=\textwidth]{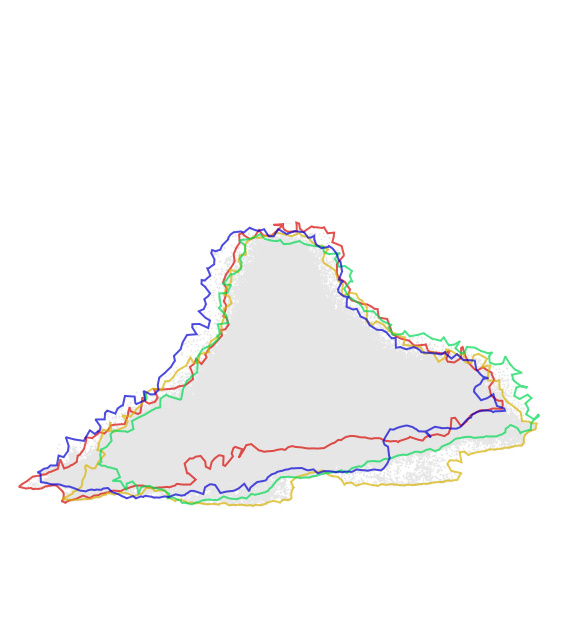}
      \caption{North-Up View.}
  \end{subfigure}
  \caption{Area covered by the ROV during the different dives.\label{fig:coverage}}
\end{figure*}

\begin{figure*}
  \centering
  \begin{subfigure}[b]{0.3\textwidth}
      \centering
      \includegraphics[width=\textwidth]{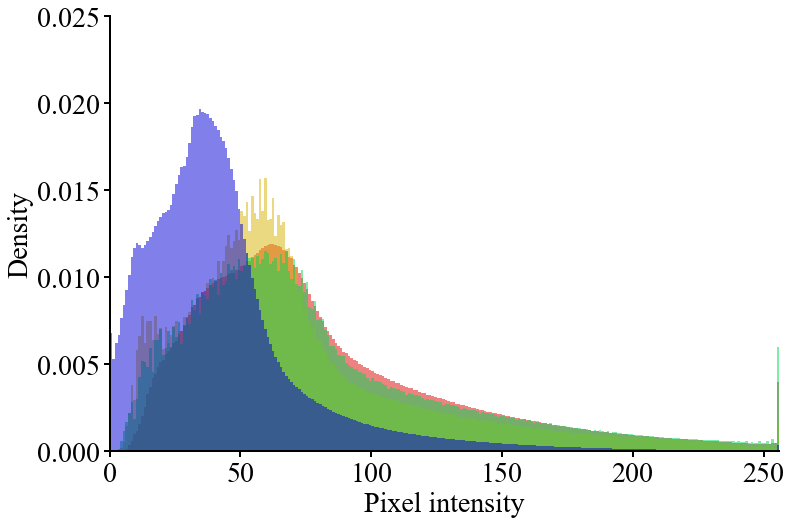}
      \caption{Red channel.}
  \end{subfigure}
  \hfill
  \begin{subfigure}[b]{0.3\textwidth}
      \centering
      \includegraphics[width=\textwidth]{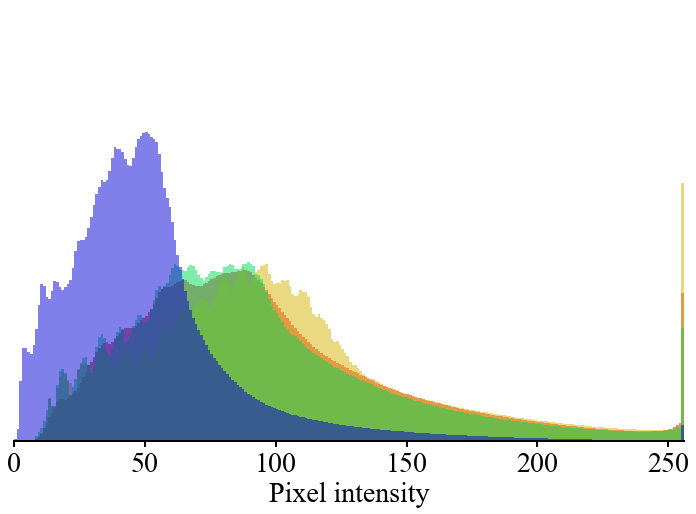}
      \caption{Green channel.}
  \end{subfigure}
  \hfill
  \begin{subfigure}[b]{0.3\textwidth}
      \centering
      \includegraphics[width=\textwidth]{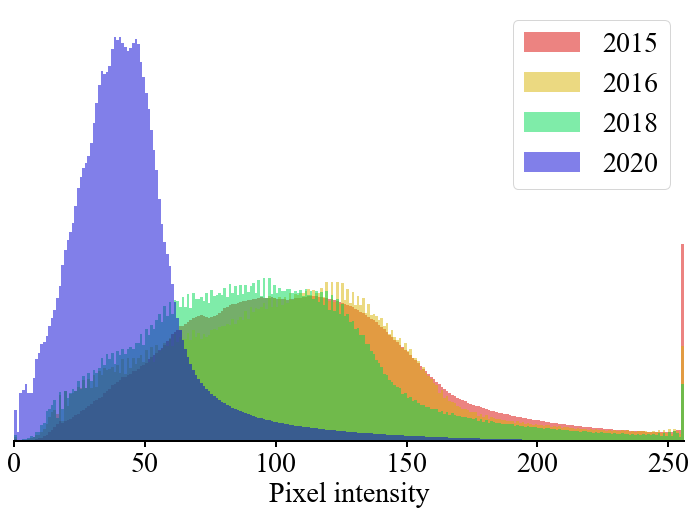}
      \caption{Blue channel.}
  \end{subfigure}
  \caption{Comparison of pixel intensity histograms for each year on each color channel.}
  \label{fig:color_hist}
\end{figure*}

This dataset contains numerous appearance changes across visits that present important challenges for the visual-based localization task. These alterations are of different nature, \textit{i.e.}, topographic, environmental or modifications in the ROV's equipment. A number of changes observed and measured over the period reflect a modification of the edifice’s local geomorphology over the years \citep{vanaudenhaege2023monitoring}. 
Chimney collapse, outcrop/boulder detachment or slide resulted in a loss of material, while vent active areas grew through mineral accretion creating new outcrops, flanges and spires. Material build-up was twice as important as the loss, suggesting that the volume of the edifice is increasing over time. While these changes can be locally drastic and affect the registration of 3D models over the years, they represent only 5\% of the total surface and are localized in areas of active venting. Local changes in hydrothermal activity also result in distinct mineralization processes, hence deposits, the color of which will vary depending on the temperature and chemical composition of the fluid.
Figure~\ref{fig:topo} reveals a modification in the topography of the scene. A chimney visible in 2015 is missing in 2020, and a temperature sensor, absent in 2015, was deployed in the vicinity in 2020.

Biological changes were more important and mainly localized in areas of topographic changes. They result from mussel populations that grow and migrate to colonize newly created habitats (143.97 m\textsuperscript{2} from 2015 to 2020) \citep{VANAUDENHAEGE2022102791}. Moreover, mussels are dynamically reoriented on a daily basis. Furthermore, the period from 2015 to 2020 showed an overall disappearance of white microbial mats over the whole edifice (-72.85 m\textsuperscript{2}). Although these changes do not affect the general topography of the structure, they strongly modify the color and texture of the model. 
Figure~\ref{fig:changes} illustrates how the mussels' population evolution over the years can alter both the topography of the scene and the colors of the vent. Also, because of these organic modifications, there are no real matching 3D points between different years on most of the chimney. This makes it difficult to match specific 2D points. Figure~\ref{fig:cross-year-matching} shows how these biological changes affect the global 3D reconstruction of the edifice. While the vent is overall well matched across different years, some specific areas like the south-east fa\c cade suffer from this source of variability and the model mostly relies on matches between images of the same year.



\begin{table*}
  \small\sf\centering
  \begin{center}
  \caption{Median localization errors and percentage of poses localized within given thresholds in meters and degrees.\label{tab:results}}
  \begin{tabularx}{\textwidth}{Xcccccccc}
      \toprule
      Method&Median errors&1 cm, 1°&2 cm, 2°&3 cm, 3°&5 cm, 5°&25 cm, 2°&50 cm, 5°&500 cm, 10°\\
      \midrule
      PoseNet & 1.98 m, 10.73° & 0.00\% & 0.00\% & 0.00\% & 0.02\% & 0.24\% & 3.58\% & 45.75\% \\
      Homoscedastic & 1.32 m, 6.27° & 0.00\% & 0.00\% & 0.00\% & 0.00\% & 0.79\% & 10.20\% & \best{62.03\%} \\
      Homography & 1.23 m, 8.30° & 0.00\% & 0.00\% & 0.00\% & 0.02\% & 0.53\% & 8.47\% & 57.83\% \\
      hLoc & 0.09 m, 1.11° & \best{15.04\%} & \best{28.37\%} & \best{36.04\%} & 43.49\% & 53.87\% & \best{57.94\%} & 60.07\% \\
      PixLoc & 6.55 m, 41.00° & 0.41\% & 1.75\% & 3.50\% & 6.49\% & 13.76\% & 15.16\% & 18.46\% \\
      hLoc+PixLoc & \best{0.08 m}, \best{1.10°} & 13.41\% & 28.18\% & 35.80\% & \best{44.08\%} & \best{53.95\%} & \best{57.94\%} & 60.07\% \\
      \bottomrule
  \end{tabularx}
  \end{center}
\end{table*}

Figure~\ref{fig:coverage} displays the area covered by the ROV each year. We notice that the vehicle covered uneven regions over the different years. The 2015 dive covered the least amount of ground compared to all other years.

Figure~\ref{fig:color_hist} compares the histograms of pixel intensities in all images of each year. First, we observe that the red channel has an overall lower pixel intensity when compared to the green and blue channels. This is easily explained by the attenuation difference of the wavelengths due to the underwater environment \citep{seathru,Berman_restoration}. We also notice a shift in pixel intensity for the 2020 dive, which is likely due to the change of camera.

\section{Visual localization benchmark}

Train and test sets were separated based on the area covered by the ROV each year. As seen on Figure \ref{fig:coverage}, the total area covered in 2016, 2018 and 2020 contains almost all the area covered in 2015. As a result, we chose 2016, 2018 and 2020 as the train set and 2015 as the test set.

Using the aforementioned train/test split, we benchmarked the \textit{Eiffel Tower} dataset on renowned visual localization methods: PoseNet with different losses \citep{kendall2015posenet,Kendall_2017_geometric,boittiaux2022homography}, hLoc \citep{sarlin2019coarse} and PixLoc \citep{sarlin2021pixloc}.
PoseNet trains a different neural network for each scene, while hLoc and PixLoc rely on deep-learning based features trained on terrestrial datasets. We detail below the parameters used for each of the methods.
\paragraph{PoseNet}\footnote{An implementation of PoseNet with all three pose regression losses is available at \href{https://github.com/clementinboittiaux/homography-loss-function}{github.com/clementinboittiaux/homography-loss-function}.}: The network as described in \citep{kendall2015posenet} is re-implemented, except for replacing the GoogLeNet backbone with a more modern MobileNetV2 \citep{mobilenetv2}. For PoseNet loss, we used $\beta = 500$ as suggested in \citep{kendall2015posenet} for the outdoor Cambridge dataset. We initialized the Homoscedastic loss as suggested in \citep{Kendall_2017_geometric}, \textit{i.e.}, $\hat{s}_x=0.0$ and $\hat{s}_q=-3.0$. For the Homography loss, we selected local $x_{min}$ and $x_{max}$ as the $2.5th$ and $97.5th$ percentile, as presented in \citep{boittiaux2022homography}.
\paragraph{hLoc}: We use the pipeline presented in \citep{sarlin2019coarse}, \textit{i.e.}, NetVLAD for image retrieval and SuperPoint alongside SuperGlue pre-trained on outdoor scenes for local matching.
\paragraph{PixLoc}: We used weights of the network pre-trained on the MegaDepth dataset \citep{MegaDepthLi18}.
\paragraph{hLoc+PixLoc}: The pose of the camera is first retrieved using hLoc and then refined with PixLoc. This pipeline is presented by \cite{sarlin2021pixloc}.

Results on the dataset for all aforementioned methods are reported in Table \ref{tab:results} and can be used as a baseline for the comparison of other long-term visual localization methods.
Unlike hLoc and Pixloc, PoseNet based methods are end-to-end networks. Consistent with the results presented by \cite{Sattler_2019_CVPR}, end-to-end networks obtain the least accurate pose estimates. Moreover, since the ground truth was constructed using SfM, methods that replicate this mode of operation, \textit{e.g.} hLoc, have an advantage because they optimize the same metric \citep{brachmann2021limits}.

hLoc and PixLoc are based on networks trained on terrestrial data, and we can expect better results by training these networks on aquatic data. However, this remains a challenge because the amount of data needed exceeds what is readily available for the underwater environment. For example, NetVLAD is trained on Google Street View Time Machine. Another approach would be to minimize the changes due to physical phenomena induced by the underwater environment to get closer to terrestrial images, using for example algorithms like \textit{Sea-thru} \citep{seathru} or SUCRe \citep{boittiaux2023sucre}.

\section{Conclusion}
This paper presented a novel dataset to evaluate visual localization methods in deep-sea environments. Unlike pre-existing datasets, \textit{Eiffel Tower} presents long-term changes in underwater scenarios, \textit{e.g.}, topography, population and species distribution, backscatter and color attenuation. We analyzed these changes and evaluated several localization pipelines on the proposed dataset. The obtained results can be used as a baseline for future work on underwater visual localization systems. Besides its use for visual localization, this dataset can also be employed to detect changes in the scene's geometry in deep-sea environments. More generally, it may also be useful to study the effects of water on various computer vision algorithms.

\begin{acks}
All data acquisitions were conducted by Ifremer. The authors would like to thank the crews of the research vessels, \textit{Pourquoi Pas?} and \textit{L'Atalante}, the pilots of the ROV Victor6000, as well as all the personnel who helped in acquiring these data. MM and LVA were supported by the European Union’s Horizon 2020 research and innovation project iAtlantic under Grant Agreement No. 818123.
\end{acks}

\bibliographystyle{SageH}

\end{document}